\documentclass[sigconf,nonacm]{acmart}

\usepackage{multicol}
\usepackage{pgfplots}

\AtBeginDocument{%
  \providecommand\BibTeX{{%
    \normalfont B\kern-0.5em{\scshape i\kern-0.25em b}\kern-0.8em\TeX}}}

\setcopyright{none}

\begin{document}

\title{Collaborative Storytelling with Large-scale Neural \\ Language Models
}

\author{Eric Nichols}
\affiliation{%
  \institution{Honda Research Institute Japan}}
\email{e.nichols@jp.honda-ri.com}
\authornote{Equal contribution.}

\author{Leo Gao}
\affiliation{%
  \institution{OrientExpress Technologies Inc.}}
\email{leogao31@gmail.com}
\authornotemark[1]

\author{Randy Gomez}
\affiliation{%
  \institution{Honda Research Institute Japan}}
\email{r.gomez@jp.honda-ri.com}

\renewcommand{\shortauthors}{Nichols et al.}
\renewcommand{\shorttitle}{Collaborative Storytelling with Large-scale Neural LMs}

\begin{abstract}
  Storytelling plays a central role in human socializing and entertainment. However, much of the research on automatic storytelling generation assumes that stories will be generated by an agent without any human interaction. In this paper, we introduce the task of {\it collaborative storytelling}, where an artificial intelligence agent and a person collaborate to create a unique story by taking turns adding to it. We present a collaborative storytelling system which works with a human storyteller to create a story by generating new utterances based on the story so far. We constructed the storytelling system by tuning a publicly-available large scale language model on a dataset of writing prompts and their accompanying fictional works. We identify generating sufficiently human-like utterances to be an important technical issue and propose a sample-and-rank approach to improve utterance quality. Quantitative evaluation shows that our approach outperforms a baseline, and we present  qualitative evaluation of our system's capabilities.
\end{abstract}




\maketitle

\section{Introduction}


Storytelling is a central part of human socialization and entertainment. Many of the popular forms of storytelling throughout history \---such as novels, plays, television, and movies\--- have passive audience experiences. However, gaming is an interesting medium because interactivity is a large part of the entertainment experience, and interactivity and storytelling can often be in conflict: too much player freedom means a storyline may never be explored, while on the other hand, too many restrictions on player freedom risks reducing gaming to a passive medium. Thus, interactivity in storytelling has been an important challenge for gaming, with much design effort put into striking a balance between entertaining gameplay and compelling storytelling.

As gaming technology advances, new opportunities for interactive storytelling present themselves. Better storage technology made telling longer, more intricate stories possible, and better graphical capabilities helped foster more immersive gaming experiences. Advances in artificial intelligence have lead to more challenging opponents, more realistic NPC behavior, and other benefits. Better procedural content generation algorithms help ensure unique gameplay experiences that stay fresh for longer. Finally, recent breakthroughs in language modeling present a new opportunity: language, and thus stories, can potentially be generated on demand. 

In this paper, we introduce a novel game of {\it collaborative storytelling}, where a human player and an artificial intelligence agent construct a story together. The game starts with the AI agent reciting one of a curated set of {\it story starters} \---opening sentences meant to kick-start participants' storytelling creativity\--- and the human player responds by adding a line, which we refer to from here on out as a {\it story continuation}, to the story. The AI agent and human player then take turns adding continuations to the story until the human player concludes the story. The game is designed to have a few restrictions as possible and contrasts with traditional storytelling settings where the narrative is fixed in advance.

\begin{figure}
    \centering
    \includegraphics[width=\columnwidth]{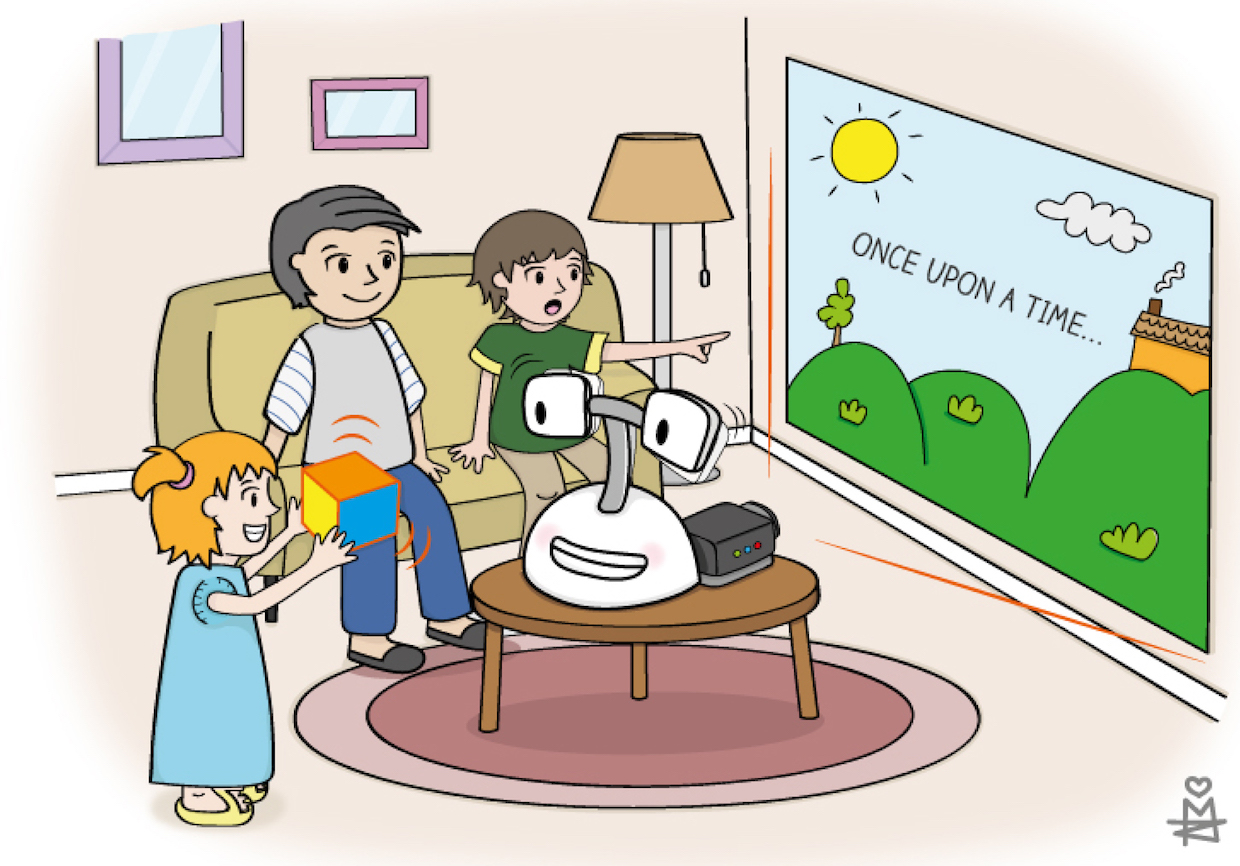}
    \caption{Collaborative storytelling with an AI agent.}
    \label{fig:haru_storytelling}
\end{figure}

Collaborative storytelling builds on a rich tradition of collaboration in storytelling that includes Dungeons and Dragons, improvisational comedy, and theater. It could be a useful tool for encouraging creativity and overcoming writer's block, as well as being an entertaining game in its own right.

Our end goal is to make it possible for intelligent agents, such as robot companions and avatars \cite{park2019model,icra20haru}, to play the collaborative storytelling game, as shown in Figure~\ref{fig:haru_storytelling}. 

Our primary contributions are as follows:

\begin{itemize}
    \item We introduce a novel task of {\it collaborative storytelling}, where humans and AI agents work together to create a story.
    \item We present a collaborative storytelling system that is constructed by tuning a large-scale neural language model on a writing prompts story dataset.
    \item We develop a method for ranking language model output to obtain more human-like story continuations.
    \item We conduct quantitative and qualitative analysis of the storytelling capabilities of our system through collaborative storytelling with human participants.
\end{itemize}

\section{Related Research}

In this section, we summarize relevant research in story generation, interactive language generation, and language modeling.

\subsection{Story Generation}


In recent years, the task of automatic story generation has gained a lot of attention. \cite{fan2018hierarchical} construct a corpus of stories and propose a hierarchical story generation model. \cite{yao2019plan} approach the task by first generating a plot outline and then filling in the language. \cite{gupta-etal-2019-writerforcing} generate story endings by incorporating keywords and context into a sequence-to-sequence model. \cite{luo2019learning} incorporate sentiment analysis into story ending generation. \cite{see2019massively} conduct an in-depth analysis of the storytelling capabilities of large-scale neural language models. However, the primary assumption of these works is that story generation is conducted without any interaction from humans.

\subsection{Interactive Language Generation}

While research dedicated to interactive language generation games is still sparse, there are a few notable recent developments. 


AI Dungeon\footnote{\url{https://play.aidungeon.io}} is a text adventure game that is generated by a GPT-2 language model \cite{radford2019language} tuned on a collection of text adventure play-throughs. In the game, players assume the first person and interact with the world by inputting commands or actions. The language model is used to generate the world's reaction to the player's actions. Our {\it collaborative storytelling} task and approach are similar to AI Dungeon, but our task is not constrained to the genre of first-person adventures, and we rank model output.


\cite{cho2020grounding} build an improvisational theater chatbot by identifying and collecting instances of improvisational dialogue on the Web and using it to tune and evaluate public domain dialogue systems. Our collaborative storytelling task is similar to improv, but stories are linguistically different enough from improv that it would be impractical to apply their dataset to our task. In addition, our approach employs sampling and ranking to improve the likelihood that language model utterances are in the desired storytelling domain, while \cite{cho2020grounding} use the model's output as-is.


\begin{figure}
    \centering
    \includegraphics[width=\columnwidth]{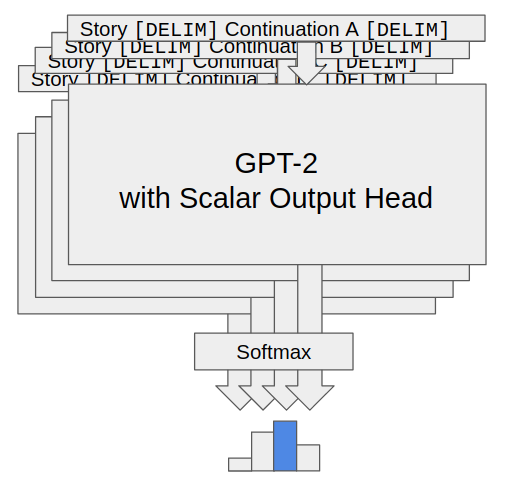}
    \caption{The ranking system architecture.}
    \label{fig:architecture}
\end{figure}

\begin{table*}
\large

\begin{tabular}{lp{1.85\columnwidth}}
\multicolumn{2}{c}{\bf Collaborative Storytelling Story 1} \\
\hline
Prompt: & {\it A man sits alone at a table, listening to the low and pleasant hum of a bassist pluck something smooth and jazzy nearby.} \\
Human: & {\it He lets out a big sigh and dreamily looks at the bassist, sipping on his whisky.} \\
System: & {\it The music is not too loud, just enough to drown out his worries, but still enough to cause him to feel like he has heard it before.} \\
Human: & {\it The man tries his best to just immerse himself into the music until it ended.} \\
System: & {\it But then the phone rang, making the man jump out of his chair.} \\
Human: & {\it "Darn it", says the man as he frantically searches his pockets for the phone.} \\ 
System: & {\it As the man was getting ready to answer, a beautiful lady, with long flowing blonde hair, greeted him.} \\ 
Human: & {\it The man drowned out the constant ringing on the phone and silenced it, saying hello back to the lady.} \\
System: & {\it She turns to the guy and smiles while staring him straight in the eyes.} \\
Human: & {\it The man replied "My name is Jacob... how about you"?} \\
System: & {\it After a few moments of silence, she smiled and explained that the man had already lived with her for years.} \\ 
Human: & {\it "Huh... what do you mean I already live with you?" Jacob said in utter confusion.} \\
System: & {\it She explained that she was the same age as he, and is the love of his life.} \\ 
Human: & {\it "No, you got to be kidding me I'm single and I come to this bar every day", Jacob said with a quiver in his voice.} \\
System: & {\it "You know what's funny?"} \\
Human: & {\it "Please explain, and are you just a stranger playing a joke on me?"} \\
System: & {\it She handed him a picture of her and a girl.} \\
Human: & {\it Her thumb was covering another person, and she slowly uncovered it as Jacob looked at the picture.} \\
System: & {\it "Look at me, " said the girl.} \\
Human: & {\it "This... picture shows me, you, and a child... but why?" Jacob started to tear up.} \\
System: & {\it "Just a moment, it will all make sense soon enough."} \\
\\
\multicolumn{2}{c}{\bf Collaborative Storytelling Story 2} \\
\hline
Prompt: & {\it The garage door lurched open and my face turned from joy to agonizing surprise.} \\
Human: & {\it A small squirrel leaped from behind my car, likely as shocked to see me as I was him.} \\
System: & {\it "I'm so sorry, I didn't mean to scare you."} \\
Human: & {\it He tilted his head to the side, almost indicating he understood my words.} \\
System: & {\it "You don't scare me."} \\
Human: & {\it Completely floored that this squirrel can speak English, I fumbled myself into asking "Are you real? This can't be real!"} \\
System: & {\it "Yes, this is real, and I have been waiting for you for quite some time."} \\
Human: & {\it I asked him "What have you been waiting for? Some peanuts from the kitchen?!"} \\
System: & {\it "No, I can answer all your questions with just one sentence!"} \\
Human: & {\it He leaned against the tire of my car with a sly look on his face, "I'm madly in love with Penny, your cat, and I'd like to take her to dinner tonight!"} \\
System: & {\it I sat there confused for a good thirty seconds before I snapped out of it.} \\
Human: & {\it "Have you been spying on my house cat!?"} \\
System: & {\it The squirrel cocked his head to the side and shrugged, as if he had no response to what I had just said.} \\
Human: & {\it "Well I'll have you know that I am not okay with you romantically pursuing my poor Penny!"} \\
System: & {\it This time, the squirrel cracked his knuckles and the look in his eyes made me question everything I was hearing.} \\
Human: & {\it "Look Mr, all I want to do here is show her a good time and maybe be a part of the family!"} \\
System: & {\it I was dumbfounded.} \\
Human: & {\it I felt a slight lump in my throat realizing that maybe this squirrel was lonely and just wanted a good home and asked "Well, I guess if you insist, where did you want to take her to dinner?"} \\
System: & {\it "To my favorite fishing spot!"} \\
Human: & {\it The squirrel chuckled, pulled out a teeny tiny fishing rod, slapped on a little fishing hat and started making his way inside to ask my soon to be very confused cat out on a date.} \\
System: & {\it Thank you for reading!} \\
\\
\end{tabular}

\caption{Example stories generated by the {\tt tuned} system with a human through the collaborative storytelling annotation task.}
\label{tbl:collaborative_stories}
\end{table*}

\subsection{Language Models}


In order for an AI agent to participate in collaborative storytelling, it must be able to generate story continuations. A language model (LM) is a mathematical model that assigns likelihoods to sequences of words where sequences that are more likely in a target language are given higher scores. Such a model can be used to generate text.

More precisely, a language model provides a probability distribution $P(x)$ over all sequences of tokens $x$. Sampling from the LM distribution is equivalent to generating text, motivating the approximation of the value of $P(x)$ using a universal function approximator, like an artificial neural network. Specifically, autoregressive LMs predict the next token given all previous tokens; this is equivalent to factoring the probability $P(x)$ as as a product of conditional probabilities of each token $x_i$ given previous tokens $x_{<i}$:

$$P(x) = \prod_{i=0}^{n} P(x_i|x_{<i})$$

The autoregressive formulation facilitates text generation as previous tokens are used to predict a distribution over potential next tokens, from which a token can be sampled and appended to the sequence. In other words, at each timestep an autoregressive LM predicts the next token given the sequence of previous tokens.

Early language models estimated token sequence likelihood based on token sequence counts taken from large collections of text together with various smoothing methods to handle novel token sequences \cite{ney1994structuring}. Later, RNNs and other sequential neural networks models became popular due to their ability to apply distributed word representations \cite{bengio2003neural,mikolov2011extensions,sutskever2011generating}, but RNNs have issues with vanishing gradients and modelling  long-term dependencies found in text.

The recent transformer architecture \cite{vaswani2017attention} uses attention layers to model long-term dependencies by greatly increasing the model's visible context. Transformers have been shown to perform well in a variety of tasks, including machine translation \cite{vaswani2017attention} and a variety of language understanding \cite{radford2019language} and language generation tasks \cite{zhang2019dialogpt}. A notable transformer model is BERT \cite{devlin2018bert}. However, as it is a bidirectional model, BERT and its variants are rarely used for text generation, due to the necessity for computationally-expensive Gibbs sampling \cite{wang2019bert}.

The model we use as a basis for our system, GPT-2 \cite{radford2019language}, is a large-scale neural network using the transformer architecture, and it has been applied successfully in a variety of language generation tasks ranging from news article generation to dialog. GPT-2 is a general purpose auto-regressive LM trained on a large corpus of internet text and its pretraining has been shown to be effective for transfer learning to novel domains.

\begin{figure*}
    \centering
    \includegraphics[width=2.1\columnwidth]{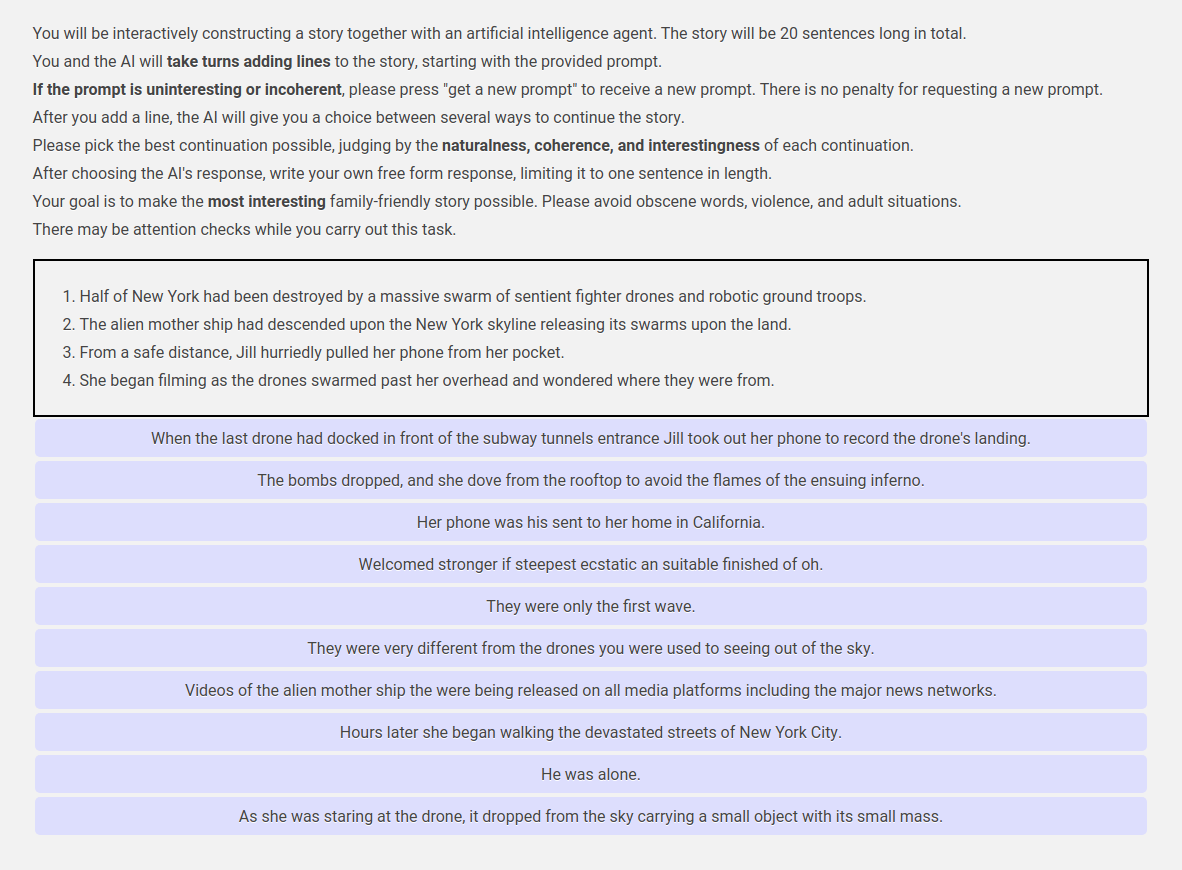}
    \caption{Web interface for collaborative storytelling annotation task. Participants select from amongst ten possible story continuations generated by the system before adding their own line to the story.}
    \label{fig:collaborative_storytelling_interface}
\end{figure*}

\section{Approach}


Our approach to collaborative storytelling is simple: a {\it Generator model} that is a large-scale neural language model tuned on storytelling data to generate story continuation candidates is combined with a {\it Ranking model} that is trained on human storyteller preferences to score them and select the highest quality continuation.

\subsection{Generation}

The Generator is a unidirectional autoregressive language model which is sampled from multiple times to generate candidate story continuations. We used the publicly-available pretrained 774M parameter GPT-2-large model\footnote{\url{https://github.com/openai/gpt-2}} tuned on our WritingPrompts dataset.

One issue with using an LM for generation is the output may be ill-formed or lacking in logical coherence. The main solutions for this issue are the use of larger models, the use of different sampling methods, and the use of various methods of traversing the search space of possible sentences. However, larger models are at greater risk of over-fitting and result in large increases in memory usage for modest gains in quality, which makes them impractical to use. As such, we focused on sampling and searching through ranking. 

\subsection{Sampling}

The most popular approaches for sampling from autoregressive models have predominantly focused on techniques for truncating the low-quality tail of the model distribution, like top-k and nucleus sampling \cite{holtzman2019curious}. Sampling is used in most GPT-2 based text generation systems, superseding greedy or untruncated sampling. In all experiments, we use nucleus sampling with $p = 0.9$. 

\subsection{Ranking}

The Ranker model scores each story continuation candidate and selects the highest scoring one. It is a standard GPT-2-large model with a final classification head consisting of a linear layer outputting a single scalar for each token. The input format to the model is: \\ {\tt (context)<|endoftext|>(choice)<|endoftext|>}.

The {\tt <|endoftext|>} token is used because it is guaranteed not to occur elsewhere in the input. As GPT-2 is unidirectional, the embedding of the final token integrates information from the entire input context window; this is similar to the use of the {\tt [CLS]} token in BERT. Thus we execute the Ranker model once for each choice, keep only the outputs from the last token of the final layer for each choice as the logit score of each choice, and compute a softmax over them. The Ranking model architecture is shown in Figure~\ref{fig:architecture}.

We chose a neural network-based Ranker model to select the best story completion from the Generator output because it offers us control over the trade-off between text generation quality and computational demand, while avoiding the significantly increased memory footprint and inflexibility in computational cost of using a larger language model. The amount of computational resources used is easily adjustable by changing the number of rollouts considered by the Ranker. This serves as a middle ground between the intractable extreme of searching the entire space of all $\mathrm{vocab}^{\mathrm{length}}$ 
possible sentences, and the computation-efficient but suboptimal solution of sampling without any branching or backtracking.

One popular alternative search solution making a similar trade-off is beam search, which keeps a dynamic list of generation candidates. Beam search has been applied in many language generation tasks, including machine translation \cite{tillmann2003word}. However, sampling from an LM using beam search can lead to degenerate text (which is typically repetitive and uninteresting), in an open-ended task such as storytelling. \cite{holtzman2019curious} These issues are avoided using a neural network-based Ranker model because it has richer text representations, it scores full text utterances rather than incomplete text fragments, and it can incorporate additional information about the storytelling domain from its training data.

\begin{table*}
    \centering
    
    \begin{multicols}{2}
    \begin{tabular}{l|l|lr}
        {\bf System} & {\bf Dataset}  & \multicolumn{2}{c}{\bf Accuracy} \\
        \hline
        {\tt tuned+ranked} & validation   & 22.9\%  & (229 / 1000) \\
        {\tt tuned+ranked} & test         & 23.3\%  & (233 / 1000) \\
        \hline \hline
        {\it random baseline} & \-- & 10.0\%  \\ 
        \hline
    \end{tabular}
    \caption{Accuracy of the {\tt tuned+ranked} model at predicting the story continuation that was selected by the Mechanical Turker who constructed the story. Note that a random baseline would pick the correct continuation 1 out of 10 times.}
    \label{tab:prediction}
    
    \begin{tabular}{r|lr}
        {\bf System}       & \multicolumn{2}{c}{\bf Acceptability} \\
        \hline
        {\tt untuned}      &     33.9\% & (305 / 900) \\
        {\tt tuned}        &     39.8\% & (358 / 900) \\
        {\tt tuned+ranker} & {\bf 62\%} & ( 62 / 100) \\ 
        \hline
    \end{tabular}
    \caption{Mean acceptability of story continuations in the test set. To evaluate {\tt untuned} and {\tt tuned}, acceptability is calculated over all 9 continuations from each system, while {\tt tuned+ranked} uses the Ranker to consider only the best one.}
    \label{tab:acceptability}
    \end{multicols}
    
\end{table*}

\subsection{Datasets}

In this section we describe our datasets: (i) a collaborative storytelling dataset constructed by crowdsourcing workers interacting with our collaborative storytelling system that are used to train the Ranker model and for evaluation, and (ii) a writing prompts dataset comprised of short stories  written in response to writing prompts posted to a Web forum that are used to train the Generator model. 

\subsubsection{Collaborative Storytelling Dataset}


We collected collaborative stories using Mechanical Turk, each consisting of 20 interactions in response to a provided story starter (which is sampled from the initial sentences of stories in the WritingPrompts dataset described in Section \ref{wpdataset}). The interactions in the story alternate between {\it choice} type interactions, in which a human participant chooses from 10 story continuations that are generated by out collaborative storytelling system, and {\it freeform} type interactions, in which the human participant is able to provide a complete sentence response. The Web interface for this task in shown in Figure~\ref{fig:collaborative_storytelling_interface}.

In order to ensure data quality, one of the continuations in the {\it choice} type interaction is a {\it distractor} which is made by concatenating randomly sampled words. The distractors are also filtered through Mechanical Turk beforehand by asking workers whether the sentences are coherent or not, and only the ones labelled incoherent by workers are used. As a quality check, if a worker selects a distractor during a {\it choice} type interaction, the story is discarded.

We collected a total of 2,200 stories, which we randomly partitioned into a training split of 2,000 stories, and validation and test splits of 100 stories each. Some example stories generated by human participants together with our system are shown in Table~\ref{tbl:collaborative_stories}.

\subsubsection{Writing Prompts Dataset}
\label{wpdataset}


We constructed a dataset of stories from the {\tt r/WritingPrompts} subreddit\footnote{\url{https://www.reddit.com/r/WritingPrompts/}}, consisting of all posts with score greater than 3 made before 2019-11-24, amounting to 140k stories in total. Some heuristics were used to clean the stories\footnote{We removed smart quotes, links and user/subreddit mentions, and all HTML entities and markdown formatting.}. This data was used to train the Generator model. 

To train the Ranker model, stories with less than 100 characters or 35 sentences were also removed. This data is then used to generate synthetic collaborative storytelling data. The first sentence of the story is used as the story starter, and the next 20 sentences are all used as the preferred story continuations of {\it choice} type interactions, where the other 9 incorrect choices are sampled from the 25th and subsequent sentences of the story.

We chose to collect our own WritingPrompts dataset instead of using the FAIR WritingPrompts dataset \cite{fan2018hierarchical}, because it gave us the flexibility to filter stories by custom score thresholds, as well as to perform the different preprocessing necessary for GPT-2. Our dataset also contains more than an additional year's worth of data compared to the FAIR dataset.

\subsection{Story Continuation Sampling and Ranking}


To generate story continuations from our system, sentences are generated from the Generator model and filtered using a set of cleanliness heuristics until the desired number of samples is achieved. Our heuristic rejected sentences with less than 60\%~alphabetic characters, unbalanced quotations, select profanity, or words like ``chapter'' that are not typically part of the story. 

For systems using ranking, the Ranker model computes a score for each story continuation and selects the highest scoring one.

\subsection{Training}

The Generator model is trained with a maximum likelihood estimation loss function using Adafactor \cite{shazeer2018adafactor} with a learning rate of 5e-5 on a weighted mixture of the WritingPrompts and BookCorpus \cite{zhu2015aligning} datasets. The addition of BookCorpus helps reduce the risk of over-fitting on the comparatively smaller WritingPrompts dataset.

The Ranking model is trained using Adam \cite{kingma2014adam} with a maximum learning rate of 1e-5. The entire model is trained; no layers are frozen. The checkpoint is resumed from a GPT-2 text generation model that was tuned on the BookCorpus and WritingPrompts datasets in the same way as the Generator model.

The Ranking model is trained on the WritingPrompts dataset and 8 copies of the training split of the Collaborative Storytelling dataset, shuffled at the story level. Each batch for the Ranking model consists of 20 sentences taken from a single story. To ensure that the model fits in memory, only the sentences that fit within 400 tokens are used, resulting in some batches with less than 20 sentences. The majority of stories do not have to be truncated.

\begin{figure*}
    \centering
    \includegraphics[width=2.1\columnwidth]{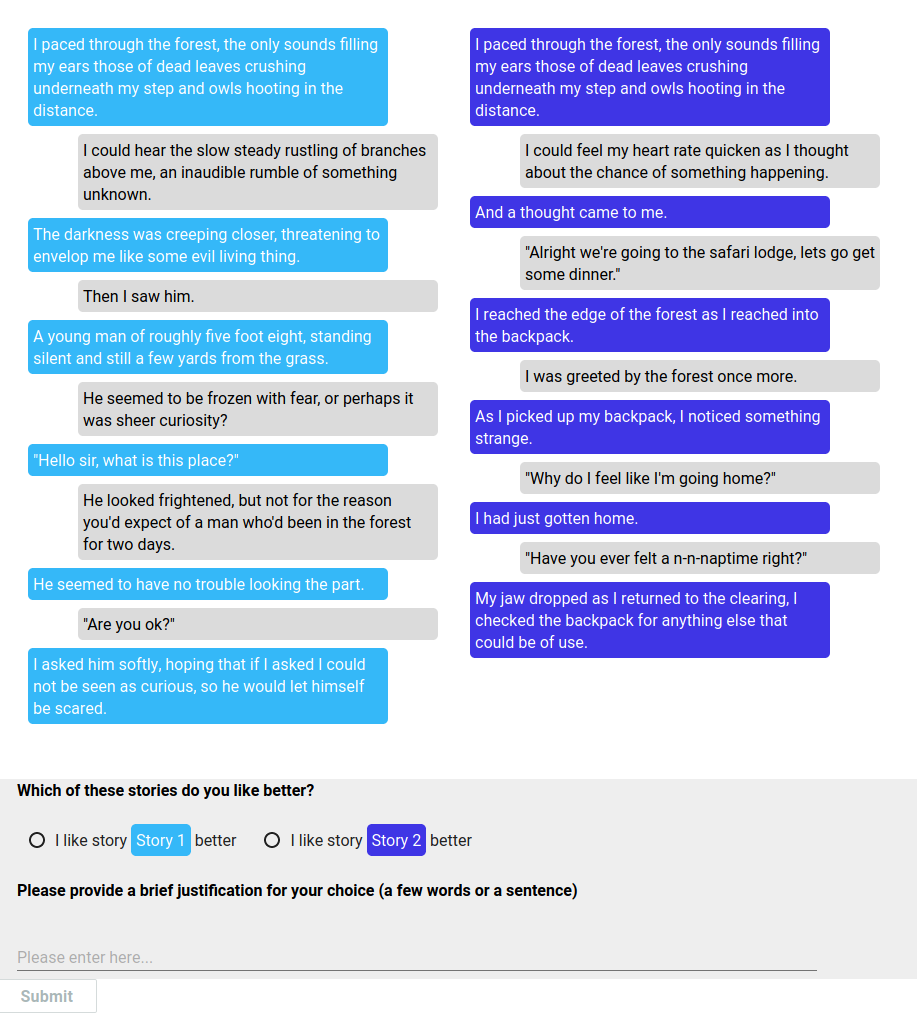}
    \caption{Web interface for storytelling system preference evaluation. }
    \label{fig:web_interface}
\end{figure*}

\begin{table*}[t]

\begin{tabular}{ll}
    {\bf Characteristic} & {\bf Question} \\
    \hline
    Engagingness & Who would you prefer to collaborate with for a long story? \\ 
    Interestingness & If you had to say one of these storytellers is interesting and one is boring, who would you say is more interesting? \\
    Humanness & Which storyteller sounds more human? \\
    Story Preference & Which of these stories do you like better? \\
    \hline
\end{tabular}
\caption{Questions asked to human evaluators of collaborative storytelling systems. Characteristics and questions are based on the {\it PersonaChat} evaluation metric of \cite{li2019acute}, with minor changes to wording to reflect the task's storytelling nature.}
\label{tab:acute_eval}

\end{table*}

\begin{figure*}
    \centering

\colorlet{xcolorA}{blue!30}
\colorlet{xcolorB}{red!30}
\colorlet{ycolorA}{red!30}
\colorlet{ycolorB}{green!30}

\begin{tikzpicture}

  \pgfplotsset{
    ybar stacked,
    axis on top,
    height=8cm, width=15.5cm,
    ymajorgrids, tick align=inside,
    major grid style={draw=white},
    enlarge y limits={value=.1,upper},
    ymin=0, ymax=100,
    ylabel={\# of participants},
    axis x line*=bottom,
    ylabel near ticks,
    ytick={0,10,20,30,40,50,60,70,80,90,100},
    yticklabels={0,,,,,50,,,,,100},
    yticklabel pos=right,
    y axis line style={opacity=0},
    tickwidth=0pt,
    enlarge x limits=true,
    xtick=data,
    symbolic x coords={
        Engagingness,
        Interestingness,
        Humanness,
        Story Preference
    },
    nodes near coords={\pgfmathprintnumber[precision=0]{\pgfplotspointmeta}},
    point meta = explicit symbolic,
    legend style={at={(0.5,0.975)},anchor=center,legend columns=3,fill=none,draw=none,column sep=0.1cm,cells={anchor=west}},
    }

  \begin{axis}[
      bar shift=-12pt,
      every node near coord/.append style={xshift=-12pt,yshift=-12pt,white},
      legend entries={untuned, tuned, tuned+ranked}
      ]
      
   \addplot+ [ybar, fill=xcolorA, draw=none, bar width=0.6cm] coordinates {
        (Engagingness,37) [37]
        (Interestingness,48) [48]
        (Humanness,40) [40]
        (Story Preference,43) [43]
    };

   \addplot+ [ybar, fill=xcolorB, draw=none, bar width=0.6cm] coordinates {
        (Engagingness,63) [63]
        (Interestingness,52) [52]
        (Humanness,60) [60]
        (Story Preference,57) [57]
    };

    \addplot+ [ybar, fill=ycolorB, draw=none, bar width=0.6cm] coordinates {
        (Engagingness,0) [62]
    };
    
    \legend{{\tt untuned}, {\tt tuned}, {\tt tuned+ranked}}
  \end{axis}
  
  \begin{axis}[
    bar shift = 12pt,
    every node near coord/.append style={xshift=12pt,yshift=-12pt,white},
    ]

    \addplot+ [ybar, fill=ycolorA, draw=none, bar width=0.6cm] coordinates {
        (Engagingness,35) [35]
        (Interestingness,43) [43]
        (Humanness,46) [46]
        (Story Preference,35) [35]
    };

   \addplot+ [ybar, fill=ycolorB, draw=none, bar width=0.6cm] coordinates {
        (Engagingness,65) [65]
        (Interestingness,57) [57]
        (Humanness,54) [54]
        (Story Preference,65) [65]
    };
    
  \end{axis}
  
\end{tikzpicture}

    \caption{Human evaluation of collaborative storytelling systems. We compare the pairs ({\tt untuned}, {\tt tuned}) and ({\tt tuned}, {\tt tuned+ranking}). Each bar graph shows a comparison of two different systems generating stories through self chat. A larger portion of the bar indicates that system was preferred by evaluators.}
    \label{fig:eval_results}
\end{figure*}
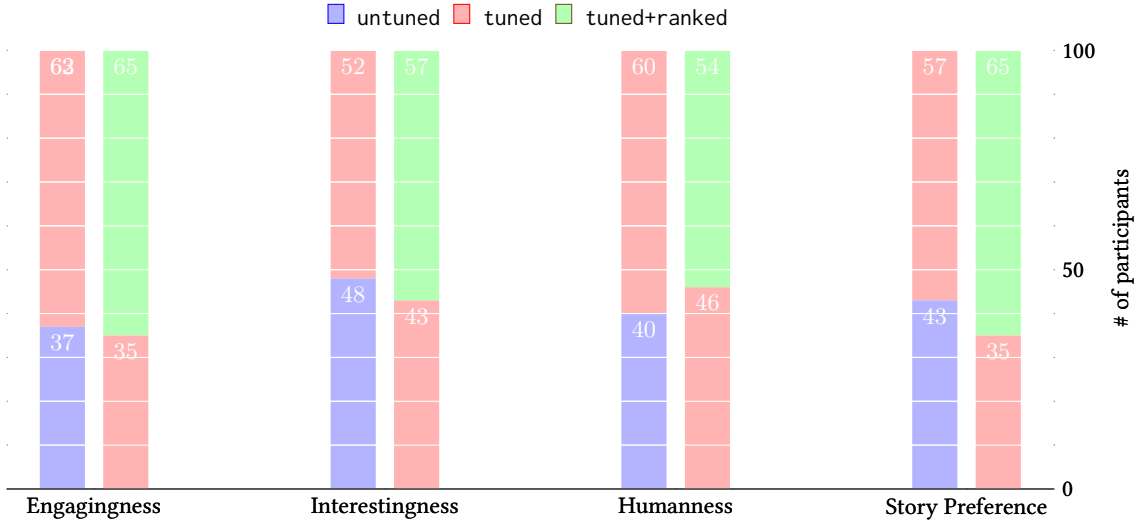

\section{Evaluation}


We evaluate our collaborative storytelling system through a combination of qualitative and quantitative metrics. To understand how well our system replicates human preferences, we measure story continuation ranking accuracy and story continuation acceptability. To gain insights into the characteristics that people feel our system has, we adapt the {\sc Acute-eval} chatbot evaluation metric \cite{li2019acute} to collaborative storytelling evaluation.

The three systems we evaluate are (i) {\tt untuned} (pretrained GPT-2) as a baseline, (ii) {\tt tuned} (GPT-2 tuned on storytelling data), and (iii) {\tt tuned+ranker} (GPT-2 tuned on storytelling data with a single story continuation selected by the Ranker model).

\subsection{Story Continuation Prediction Accuracy}


Story continuation prediction accuracy measures the accuracy of the Ranker model at predicting the continuation chosen by the Mechanical Turk worker that interacted with the model to produce the story. This metric is a proxy for how often the {\tt tuned+ranked} picks the best continuation of the story, but its usefulness is diminished by variance in human annotators and the possibility of multiple equally good continuations. The results are summarized in Table~\ref{tab:prediction}. Nonetheless, we find that our Ranker model outperforms chance by a factor of over two, providing evidence that it is able to capture the preferences of human annotators to an extent.

\begin{table*}
\large

\begin{tabular}{lp{1.85\columnwidth}}
\multicolumn{2}{c}{\bf Self-chat tuned+ranked Story 1} \\
\hline
A: & {\it The clock finally struck 10am and the massive archway gates were slowly pulled open with the sound of a grinding mechanism.} \\
B: & {\it I stood before the immense bronze doors, arms crossed, awaiting the fate of the two evil Dictators. } \\
A: & {\it “Ladies and gentlemen of the land” a soft yet authoritative voice boomed through the doors. } \\
B: & {\it “We are here for your arrival” a commanding and officious voice echoed through the halls. } \\
A: & {\it A slow breath tickled my lips as I was finally allowed inside. } \\
B: & {\it The archway was brimming with people, many in red, with metal cufflinks, and most in ornate black garb. } \\
A: & {\it I stepped forward into the main audience hall, pausing to take in my surroundings. } \\
B: & {\it The ceiling was made of rough jagged glass panels; out in the back stood eight shiny and elaborately painted metal ladders. } \\
A: & {\it The stately mahogany wooden podium was suspended from the ceiling, very tall and proud. } \\
B: & {\it The four different Dictators circled the podium with their various groups in the room. } \\
A: & {\it The oldest of the Dictators stood, hands together in front of him, on his magnificent golden podium. } \\
B: & {\it He seemed more imposing than the rest, holding a bony silver sword by the hilt in his shining gold hand. } \\
A: & {\it “Hallowed be the name of his majesty, King Gautir” the King boomed through the open archway. } \\
B: & {\it The King was a corpulent, middle aged man with his long white hair growing out of his crown, nearly the size of his small face. } \\
A: & {\it I could see that his eyes were piercing like the flaming sun. } \\
B: & {\it The King’s long, white robes seemed long like a black robe worn by someone with a wide girth. } \\
A: & {\it His sneering smile was reflected in his large, deep blue eyes. } \\
B: & {\it “Today, we greet you both, members of the committee, from the Holy Empire of All Flame and Sound.” } \\
A: & {\it The King’s voice boomed as the rest of the Dictators turned and stepped towards the podium, stopping on their footsteps. } \\
B: & {\it “Welcome to this evening’s meeting” the King announced to us. } \\
A: & {\it “You two can either choose to remain in your chambers or come out and face the Committee” The King said in a commanding tone. } \\
\\
\multicolumn{2}{c}{\bf Self-chat tuned+ranked Story 2} \\
\hline
A: & {\it Have you ever wondered what exactly goes into making our favorite foods?} \\
B: & {\it You know, the big baked potato with oil and salt, the slushy that's savory enough to eat for lunch every day? } \\
A: & {\it The answer is simple: the food industry. } \\
B: & {\it It’s a fairly big industry, you see. } \\
A: & {\it It’s sort of estimated that five hundred different corporations control 70\% of the world’s food supply. } \\
B: & {\it If you were to think of it in simple terms, that’s right: almost everyone on the planet eats food produced by a food corporation. } \\
A: & {\it As you might imagine, the world is full of giant corporations and a huge amount of money being made through the food industry. } \\
B: & {\it We’ve all heard the expression “money talks, money does.” } \\
A: & {\it What do you think happens when that money goes through the food industry? } \\
B: & {\it It goes into the mouths of politicians and it goes into the mouths of stockholders. } \\
A: & {\it This whole industry is very complicated, although people never seem to come up with any clear-cut answers for it. } \\
B: & {\it But, really, the only thing that people seem to have a handle on about the food industry is that it’s a big financial mess. } \\
A: & {\it When you think about it, this seems pretty clear. } \\
B: & {\it Let me explain. } \\
A: & {\it In order to make money, companies have to produce enough food to feed a population. } \\
B: & {\it When a company creates more food than it can eat, it has to buy more food, so that it can keep producing more food. } \\
A: & {\it More food means more food prices, so that the corporation can make more money. } \\
B: & {\it So how does the food industry achieve this? } \\
A: & {\it The answer to this is pretty simple. } \\
B: & {\it In the world of food production, companies come up with marketing schemes that manipulate people's tastes. } \\
A: & {\it Using a carrot for example, a company might work to improve a persons’ reaction to carrots. } \\
\\
\end{tabular}

\caption{Example stories generated by self-chat with the {\tt tuned+ranked} system.}
\label{tbl:self_chat_stories}
\end{table*}

\subsection{Story Continuation Acceptability}


As an additional measure of our systems' capacity to generate story continuations that match human preferences, we formulate the story continuation acceptability task. In this task, each story continuation generated by a system is classified as either {\it acceptable} or {\it unacceptable}, and we compare their mean acceptability precision.

We annotated the acceptability of candidate story continuations by asking Mechanical Turk workers to classify each continuation given the context of the story generated so far. To ensure annotation quality, we have 3 workers evaluate each {\it choice} interaction per story from both the validation and test sets and take the majority vote across the three labels as the final label\footnote{The workers reached unanimous agreement 41.9\% of the time on the test data.}. These {\it choice} interactions consist of 9 story continuations generated by the system and 1 incoherent distractor. If a worker labels a distractor acceptable, their annotations are discarded. We use this method to evaluate how often each model produces outputs that are an acceptable continuation of the story, rather than the best continuation.

Since the {\tt tuned} and {\tt tuned+ranked} systems use the same language model samples, we use the test set to evaluate their performance, considering the mean acceptability of all of the sampled continuations from {\tt tuned} and the acceptability of the single continuation selected by {\tt tuned+ranked} for each {\it choice} interaction in the datasets. To evaluate the {\tt untuned} system, we gather and evaluate 100 {\it choice} interactions by having Mechanical Turkers construct stories with the {\tt untuned} system.

The results are summarized in Table~\ref{tab:acceptability}. As we can see, the {\tt tuned} system outperforms the {\tt untuned} system, showing that tuning the language model on storytelling data is important in improving generation quality. We also find that {\tt tuned+ranked} greatly outperforms the other two systems, providing supporting evidence that our Ranking model is effective at helping our language model produce story continuations that are likely to be preferred by humans. 

\subsection{Human Annotator Story Preferences}


Conducting qualitative evaluation of collaborative storytelling is challenging because the highly interactive nature of the task means that the influence of human participants makes it difficult to isolate the performance of the system. Ideally we would like to conduct subjective evaluation of participants' collaborative storytelling experience with an intelligent agent, but this is left for future work.

Instead, since collaborative storytelling involves language exchange between entities with turn taking, we take inspiration from dialogue system evaluation methodology. Faced with the challenge of comparing multiple dialogue systems, \cite{li2019acute} developed a method of comparing conversation pairs that instructs evaluators to only pay attention to the contributions of a single specified speaker in the conversation. In addition, their evaluation method, known as {\sc Acute-eval}, allowed them to evaluate the contributions of a given dialogue system in terms of characteristics, such as engagingness, interestingness, humanness, and knowledgeability. Finally, to evaluate different dialogue systems without requiring a human to chat with them, they apply the self-chat technique of \cite{ghandeharioun2019approximating} and generate conversations for evaluation by having dialogue systems talk to themselves.

We create our own evaluation metric based on the characteristics targeted by the {\it PersonaChat} metric of {\sc ACUTE-Eval}\footnote{We exclude the {\it Wizard of Wikipedia} metric because knowledgeability is not directly relevant to our collaborative storytelling setting.}. For each target characteristic, we take the question that \cite{li2019acute} identified as most likely to differentiate between the evaluation of two systems and reword it to fit the collaborative storytelling setting. Finally, we add a question to measure overall story preference. The resulting evaluation metric is shown in Table~\ref{tab:acute_eval}.

We created a Mechanical Turk task to determine relative pairwise user preferences using our evaluation metric. To eliminate variance from human storytellers, we use the self-chat setting of \cite{li2019acute}, where each model converses with itself. Some example stories are shown in Table~\ref{tbl:self_chat_stories}. We compare the {\tt untuned} and {\tt tuned+ranked} models against the {\tt tuned} model. For each pair of models, we collect 100 comparisons per question, and we instruct workers to provide short justifications for their decisions. The Web interface shown to workers is given in Figure~\ref{fig:web_interface}. 

The results of the evaluation are summarized in Figure~\ref{fig:eval_results}. For each characteristic evaluated, the pairs of models are shown as stacked bar graphs, where a larger portion represents a stronger preference for that system. As can be seen, {\tt tuned} is preferred over {\tt untuned}, and {\tt tuned+ranked} is preferred over {\tt tuned} for all characteristics and overall story preferences, providing evidence that tuning the language model on storytelling data and ranking the generated story continuations make complementary contributions to our collaborative storytelling system's performance.

\section{Discussion}


In this section, we discuss the advantages and limitations of our approach to collaborative storytelling.

\subsection{Advantages}

The advantages of our approach are that our storytelling system can produce well-formed story contributions that display creativity and react to the contributions made by human storytellers. In Collaborative Storytelling Story 1 from Table~\ref{tbl:collaborative_stories}, we see an example of that creativity, when our system introduces the plot twist that the man and women not only know each other but have been living together for year. In Story 2 from the same table, we see our system's ability to play along with a human storyteller when the system accepts its collaborator's assertion that the squirrel can speak English and starts crafting dialogue for it.

\subsection{Limitations}

The limitations of our approach are that our storytelling system has a very shallow model of the world, which can lead to incoherent output. This is illustrated by the self-chat Story 2 in Figure~\ref{fig:web_interface}: the narrative makes jarring shifts in setting and lacks overall cohesion. Such problems in cohesion are often amplified in self-chat settings, as the model lacks human input to reign it in.

In addition, because the storytelling model lacks explicit story  structure, it can be hard to steer toward desired output, such as a human-preferred genre or mood, or generation of story endings on demand. We plan to address these issues in future work by adding more structure to the data used to train our models. 

Finally, evaluation of this task is challenging: because interaction with human players introduces variance into the output, it is difficult to directly compare generated stories, but at the same time, evaluation limited to self-chat is not fully reflective of our desired task setting. Once our system has been implemented in a suitable agent, we plan to carry out detailed subjective evaluation of the collaborative storytelling experience of volunteers to gain further insights about our task and approach. 



\section{Conclusion}


In this paper, we introduced the novel task of {\it collaborative storytelling}, where humans and AI agents work together to make stories. We presented a collaborative storytelling system that tunes a large-scale neural LM on storytelling data and uses a sampling-and-ranking approach to select more human-preferred story continuations. Quantitative evaluation of our system found that tuning and ranking both greatly contribute to its capability to generate story continuations that human evaluators prefer and consider acceptable. Qualitative evaluation of human evaluator preferences showed that humans found {\tt tuned+ranked} more preferable than {\tt tuned} and {\tt tuned} more preferable than {\tt untuned} in terms of engagingness, interestingness, and humanness metrics, as well as overall story quality preferences. Finally, we identified areas for potential future work, including evaluation of stories produced by humans and our system, integration of our system into intelligent agents such as robots and avatars, and improvement of generated story continuation quality by allowing genres or moods to be targeted.

\bibliographystyle{ACM-Reference-Format}
\bibliography{main}



\end{document}